\documentclass[sigplan,screen]{acmart}

\AtBeginDocument{%
  \providecommand\BibTeX{{%
    \normalfont B\kern-0.5em{\scshape i\kern-0.25em b}\kern-0.8em\TeX}}}

\copyrightyear{2022}
\acmYear{2022}
\setcopyright{rightsretained}
\acmConference[UIST '22 Adjunct]{The Adjunct Publication of the 35th Annual ACM Symposium on User Interface Software and Technology}{October 29-November 2, 2022}{Bend, OR, USA}
\acmBooktitle{The Adjunct Publication of the 35th Annual ACM Symposium on User Interface Software and Technology (UIST '22 Adjunct), October 29-November 2, 2022, Bend, OR, USA}
\acmDOI{10.1145/3526114.3558712}
\acmISBN{978-1-4503-9321-8/22/10}

\begin{document}

\title{\textit{PerSign:} Personalized Bangladeshi Sign Letters Synthesis}


\author{Mohammad Imrul Jubair}
\affiliation{%
  \institution{University of Colorado Boulder, United States}
  \city{}
  \country{}
}
\email{mohammad.jubair@colorado.edu}

\author{Ali Ahnaf}
\affiliation{%
   \institution{Ahsanullah University of Science and Technology, Bangladesh}
  \city{}
  \country{}
 }
 \email{aliahnaf327@gmail.com}

\author{Tashfiq Nahiyan Khan}
\affiliation{%
   \institution{Ahsanullah University of Science and Technology, Bangladesh}
  \city{}
  \country{}
  }
  \email{khantashfiq565@gmail.com}

\author{Ullash Bhattacharjee}
\affiliation{%
  \institution{Ahsanullah University of Science and Technology, Bangladesh}
  \city{}
  \country{}
 }
\email{turja997@gmail.com}

\author{Tanjila Joti}
\affiliation{%
    \institution{Ahsanullah University of Science and Technology, Bangladesh}
  \city{}
  \country{}
  }
\email{tanjilajoti@gmail.com}

\renewcommand{\shortauthors}{Jubair and Ahnaf, et al.}

\begin{abstract}
Bangladeshi Sign Language (BdSL)---like other sign languages--- is tough to learn for general people, especially when it comes to expressing letters.
In this poster, we propose \textit{PerSign}, a system that can reproduce a person's image by introducing sign gestures in it. We make this operation \textit{``personalized''}, which means the generated image keeps the person's initial image profile--face, skin tone, attire, background---unchanged while altering the hand, palm, and finger positions appropriately. We use an image-to-image translation technique and build a corresponding unique dataset to accomplish the task. We believe the translated image can reduce the communication gap between signers\footnote{person who uses sign language.} and non-signers without having prior knowledge of BdSL.
\end{abstract}

\ccsdesc[300]{Human-centered computing~Text input}
\keywords{Bangladeshi Sign Language (BdSL), Image-to-Image Translation, Sign Letters Synthesis.}

\begin{teaserfigure}
  \includegraphics[width=\textwidth]{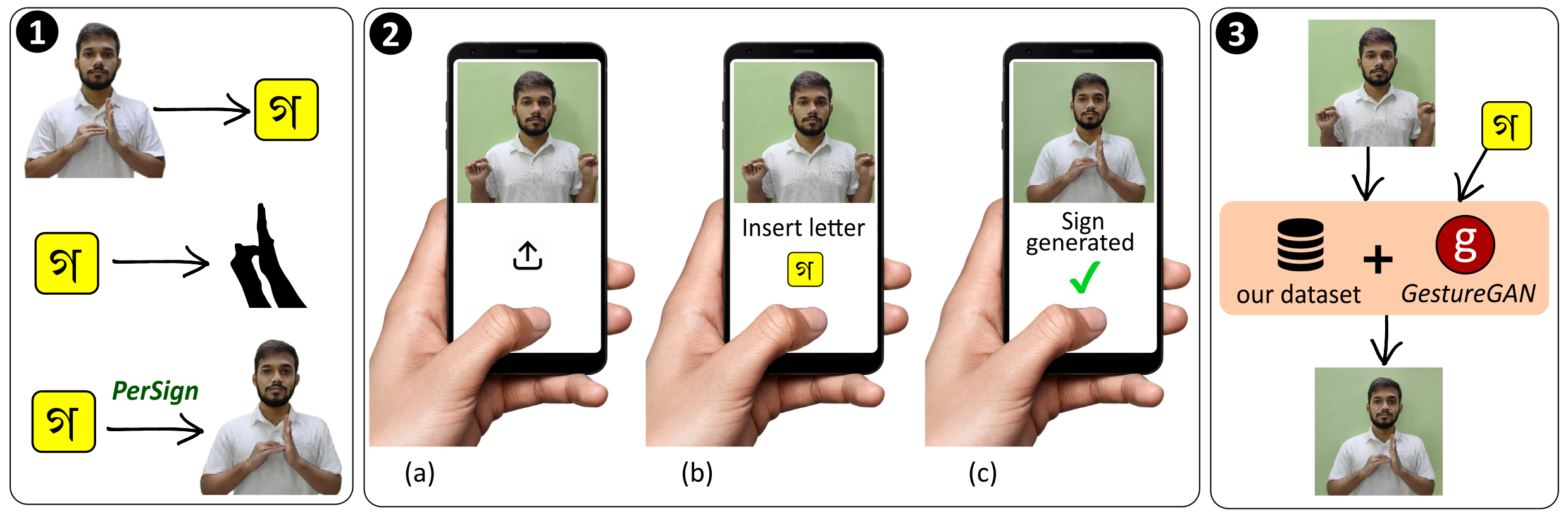}
  \caption{An overview of our \textit{\textbf{PerSign}}. Here, {\protect\includegraphics[scale=.2]{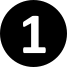}}
 shows the novelty of the work. \textit{Top row} of \protect\includegraphics[scale=.2]{crc1.png} indicates the Bangladeshi sign language (BdSL) classification where an image with gesture is classified as the Bengali letter \protect\includegraphics[scale=.17]{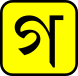}.
\textit{Middle} exhibits the vice versa of the previous one, where symbols of signs are generated using generative adversarial networks (GANs).
In contrast, ours (\textit{last}) generate personalized signs---it reproduces a person's image with the desired sign for a given letter while keeping the profile---face, dress, etc.---unchanged. \protect\includegraphics[scale=.2]{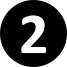} shows an abstraction of our working pipeline (see section-\ref{sec:intro}).
 \protect\includegraphics[scale=.2]{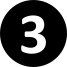} shows image translation module that takes letter and profile image, and GAN \protect\includegraphics[scale=.125]{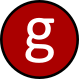} translates the image by learning from our dataset.}
 \Description{An illustration of PerSign showing the process of the application and also illustrating the overall architecture of the system}
  \label{fig:teaser}
\end{teaserfigure}

\maketitle

\section{Introduction}
\label{sec:intro}
About $13$ million people in Bangladesh are suffering from different degrees of hearing loss, of which $3$ million have hearing disability~\cite{alauddin2004deafness}.
There are around $1$ million using Bangladeshi Sign Language (BdSL) in their everyday life\cite{c2}. While communicating with a signer
, there are two major tasks for a non-signer: \textit{(i)} understanding the signs and \textit{(ii)} expressing the signs. Researchers made impressive contributions to task~\textit{(i)} by developing sign letters\footnote{signs that represent letters only.} recognition techniques from images (Fig.~\ref{fig:teaser}~\protect\includegraphics[scale=.18]{crc1.png}). Several works has been proposed for BdSL letters classification via machine learning techniques~\cite{islam2022improving, rahim2022soft, miah2022bensignnet, hasan2021shongket, khatun2021systematic, talukder2021okkhornama, Hoque_2020_ACCV, BdSLiciet}.
Task \textit{(ii)} still has less research attention since it is a difficult process for non-signers. A naive and tiresome approach to expressing signs is to use flashcards with signs and symbols. Being inspired by that,~\cite{shishir2020esharagan} proposed a system that generates symbols of signs using generative adversarial networks (GANs)~\cite{goodfellow2014generative}. However, their work only produces symbols of signs---which may raise questions regarding the necessity of such a system. Another way is by using animated avatars of signs, i.e.~\cite{KippAvat}; but there is no such system for BdSL. Moreover, an avatar-based system does not provide a realistic environment for communication.

All the above scenarios inspired us to make task~\textit{(ii)} more realistic yet effortless. Hence, we introduced \textit{\textbf{PerSign}: {Pers}onalized Bangladeshi {Sign} Letters Synthesis} which converts the image of a user into an image showing signs while keeping the person's profile unchanged. Fig.~\ref{fig:teaser}~\protect\includegraphics[scale=.19]{crc2.png} explains the working pipeline of our prototype. A user first uploads their profile photo ($I_P$) only once to our system. The $I_P$ must follow a specific rule of showing hand and palm (as shown in~\protect\includegraphics[scale=.19]{crc2.png}{$^{a}$}). After that, the user inserts the desired letter ($L$) to be expressed (e.g. \protect\includegraphics[scale=.17]{Ga.png} in \protect\includegraphics[scale=.19]{crc2.png}{$^{b}$}). Our system converts $I_P$ into $I_L$ by considering $L$. This can be seen as $I_L \leftarrow I_P+L$, where $I_L$ contains the same person in $I_P$ with unchanged face, skin tone, attire, and background (\protect\includegraphics[scale=.19]{crc2.png}{$^{c}$}). In that case, the person does not need any expertise in sign language. We believe, a signer will feel a natural environment if $I_L$ is shown, thus, making the communication more realistic and affectionate.

\textbf{\textit{\color{blue}Do we really need such a system?}}---in order to address this question, we performed a survey on a group of $6$ {guardians} and $11$ {teachers} of deaf children---who were also signers---regarding the necessity of our system.
We let participants upload profile images to \textit{PerSign} and asked them to rate the results on a scale of \textcircled{\small 1} to \textcircled{\small 5} according to \textit{Likert} rating method~\cite{likert1932technique}, with \textcircled{\small 1} being \textit{not necessary at all} and \textcircled{\small 5} being \textit{very necessary}. Out of total $17$ participants, $13$ and $3$ rated \textit{PerSign} with \textcircled{\small 5} and \textcircled{\small 4} respectively with an average rating of $4.705$. Most of the sign language teachers commented that \textit{personalized} signs are very helpful for general people to get closer to signers, especially when it comes to children.

\section{Implementation}
\begin{figure}[htb]
  \includegraphics[width=0.49\textwidth]{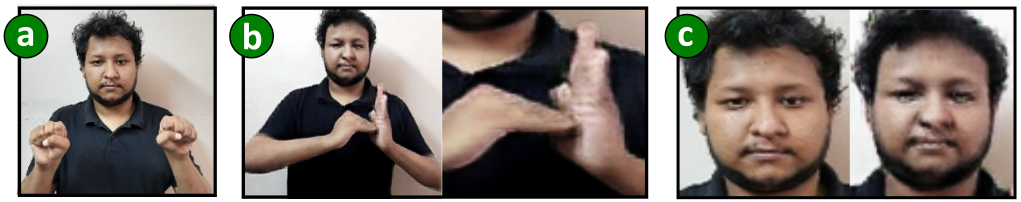}
  \caption{Result analysis. (a) input profile image. (b) generated image with signs. (c) zoomed view of faces from input \textit{(left)} and output \textit{(right)}. We can see, that the face is retained.}
  \label{fig:res}
  \Description {A picture of the result obtained exclusively showing the face and hand-gestures.}
\end{figure}

We employed the \textit{Generative Adversarial Network} (Fig.~\ref{fig:teaser}~\protect\includegraphics[scale=.18]{crc3.png}), an unsupervised deep learning technique that automatically learns the patterns from datasets in order for the model to produce a new output~\cite{goodfellow2014generative}. Our problem lies under a sub-domain of GAN which \textit{image-to-image translation}~\cite{isola2017image} and we adopted \textit{GestureGAN}~\cite{tang2018gesturegan}---a gesture-to-gesture translation method---to implement a prototype system. For this purpose, we built a dataset of images with hand gestures of arbitrary poses, sizes, and backgrounds.
Since we needed a paired dataset $\{I_P, I_L\}$ we could not reuse any of the existing unpaired ones. For localizing gestures and appearances, we exploited \textit{OpenPose}~\cite{8765346} to make the skeleton of the hands and face of the images and store the pair of input and output images. We then trained and tested the \textit{GestureGAN} model with our dataset, and constructed our system's working prototype. Fig.~\ref{fig:res} presents more results of \textit{PerSign}.

\section{Conclusion and Future work }
In this poster, we proposed a framework---\textit{PerSign}---for synthesizing Bangladeshi Sign Letters that can be \textit{personalized}. Through our method, anyone will be able to communicate as a signer without having any expertise in BdSL. We built our own dataset and exploited \textit{GestureGAN} method to accomplish the task. Our work is still in progress and we gathered comments from the participants during the survey to find areas for improvement. Most of the users recommended applying the technique for a sequence of images to render video for a stream of input letters. We can achieve better results by increasing the number of diverse examples in our dataset. Though our dataset has samples for all BdSL letters, some specific letters need to be treated carefully because of their similarities in patterns.
Our final  aim is to merge task (\textit{i})
and (\textit{ii})
into a single system to provide an \textit{one-stop solution} for two-way communication. Some of the users also suggested extending our work for full gestures, rather than letters only. We are also intent on implementing better \textit{GUI} with voice input. Last but not the least, we plan to conduct a thorough evaluation from experts and signers. We believe, this poster opens new avenues in sign language for further research.

\newpage
\newpage
\bibliographystyle{ACM-Reference-Format}
\bibliography{sample-base}


\begin{thebibliography}{17}


\ifx \showCODEN    \undefined \def \showCODEN     #1{\unskip}     \fi
\ifx \showDOI      \undefined \def \showDOI       #1{#1}\fi
\ifx \showISBNx    \undefined \def \showISBNx     #1{\unskip}     \fi
\ifx \showISBNxiii \undefined \def \showISBNxiii  #1{\unskip}     \fi
\ifx \showISSN     \undefined \def \showISSN      #1{\unskip}     \fi
\ifx \showLCCN     \undefined \def \showLCCN      #1{\unskip}     \fi
\ifx \shownote     \undefined \def \shownote      #1{#1}          \fi
\ifx \showarticletitle \undefined \def \showarticletitle #1{#1}   \fi
\ifx \showURL      \undefined \def \showURL       {\relax}        \fi
\providecommand\bibfield[2]{#2}
\providecommand\bibinfo[2]{#2}
\providecommand\natexlab[1]{#1}
\providecommand\showeprint[2][]{arXiv:#2}

\bibitem[Alauddin and Joarder(2004)]%
        {alauddin2004deafness}
\bibfield{author}{\bibinfo{person}{Mohammad Alauddin} {and}
  \bibinfo{person}{Abul~Hasnat Joarder}.} \bibinfo{year}{2004}\natexlab{}.
\newblock \showarticletitle{Deafness in bangladesh}.
\newblock In \bibinfo{booktitle}{\emph{Hearing Impairment}}.
  \bibinfo{publisher}{Springer}, \bibinfo{pages}{64--69}.
\newblock


\bibitem[{Cao} et~al\mbox{.}(2019)]%
        {8765346}
\bibfield{author}{\bibinfo{person}{Z. {Cao}}, \bibinfo{person}{G. {Hidalgo
  Martinez}}, \bibinfo{person}{T. {Simon}}, \bibinfo{person}{S. {Wei}}, {and}
  \bibinfo{person}{Y.~A. {Sheikh}}.} \bibinfo{year}{2019}\natexlab{}.
\newblock \showarticletitle{OpenPose: Realtime Multi-Person 2D Pose Estimation
  using Part Affinity Fields}.
\newblock \bibinfo{journal}{\emph{IEEE Transactions on Pattern Analysis and
  Machine Intelligence}} (\bibinfo{year}{2019}).
\newblock


\bibitem[Goodfellow et~al\mbox{.}(2014)]%
        {goodfellow2014generative}
\bibfield{author}{\bibinfo{person}{Ian Goodfellow}, \bibinfo{person}{Jean
  Pouget-Abadie}, \bibinfo{person}{Mehdi Mirza}, \bibinfo{person}{Bing Xu},
  \bibinfo{person}{David Warde-Farley}, \bibinfo{person}{Sherjil Ozair},
  \bibinfo{person}{Aaron Courville}, {and} \bibinfo{person}{Yoshua Bengio}.}
  \bibinfo{year}{2014}\natexlab{}.
\newblock \showarticletitle{Generative adversarial nets}.
\newblock \bibinfo{journal}{\emph{Advances in neural information processing
  systems}}  \bibinfo{volume}{27} (\bibinfo{year}{2014}).
\newblock


\bibitem[Hasan et~al\mbox{.}(2021)]%
        {hasan2021shongket}
\bibfield{author}{\bibinfo{person}{SK~Nahid Hasan}, \bibinfo{person}{Md~Jahid
  Hasan}, {and} \bibinfo{person}{Kazi~Saeed Alam}.}
  \bibinfo{year}{2021}\natexlab{}.
\newblock \showarticletitle{Shongket: A Comprehensive and Multipurpose Dataset
  for Bangla Sign Language Detection}. In \bibinfo{booktitle}{\emph{2021
  International Conference on Electronics, Communications and Information
  Technology (ICECIT)}}. IEEE, \bibinfo{pages}{1--4}.
\newblock


\bibitem[Hoque et~al\mbox{.}(2020)]%
        {Hoque_2020_ACCV}
\bibfield{author}{\bibinfo{person}{Oishee~Bintey Hoque},
  \bibinfo{person}{Mohammad~Imrul Jubair}, \bibinfo{person}{Al-Farabi Akash},
  {and} \bibinfo{person}{Saiful Islam}.} \bibinfo{year}{2020}\natexlab{}.
\newblock \showarticletitle{BdSL36: A Dataset for Bangladeshi Sign Letters
  Recognition}. In \bibinfo{booktitle}{\emph{Proceedings of the Asian
  Conference on Computer Vision (ACCV) Workshops}}.
\newblock


\bibitem[Hoque et~al\mbox{.}(2018)]%
        {BdSLiciet}
\bibfield{author}{\bibinfo{person}{Oishee~Bintey Hoque},
  \bibinfo{person}{Mohammad~Imrul Jubair}, \bibinfo{person}{Md.~Saiful Islam},
  \bibinfo{person}{Al-Farabi Akash}, {and} \bibinfo{person}{Alvin~Sachie
  Paulson}.} \bibinfo{year}{2018}\natexlab{}.
\newblock \showarticletitle{Real Time Bangladeshi Sign Language Detection using
  Faster R-CNN}. In \bibinfo{booktitle}{\emph{2018 International Conference on
  Innovation in Engineering and Technology (ICIET)}}. \bibinfo{pages}{1--6}.
\newblock
\urldef\tempurl%
\url{https://doi.org/10.1109/CIET.2018.8660780}
\showDOI{\tempurl}


\bibitem[Islam et~al\mbox{.}(2022)]%
        {islam2022improving}
\bibfield{author}{\bibinfo{person}{Sohal Islam}, \bibinfo{person}{Showni~Rudra
  Titli}, \bibinfo{person}{Kazi~Arham Kabir}, \bibinfo{person}{Md~Abdullah
  Al~Hossain}, {and} \bibinfo{person}{Md~Altaf Hossain}.}
  \bibinfo{year}{2022}\natexlab{}.
\newblock \showarticletitle{Improving Real-time Hand Gesture Recognition System
  for Translation: Sensor Development}. In \bibinfo{booktitle}{\emph{2022 17th
  Annual System of Systems Engineering Conference (SOSE)}}. IEEE,
  \bibinfo{pages}{254--259}.
\newblock


\bibitem[Isola et~al\mbox{.}(2017)]%
        {isola2017image}
\bibfield{author}{\bibinfo{person}{Phillip Isola}, \bibinfo{person}{Jun-Yan
  Zhu}, \bibinfo{person}{Tinghui Zhou}, {and} \bibinfo{person}{Alexei~A
  Efros}.} \bibinfo{year}{2017}\natexlab{}.
\newblock \showarticletitle{Image-to-image translation with conditional
  adversarial networks}. In \bibinfo{booktitle}{\emph{Proceedings of the IEEE
  conference on computer vision and pattern recognition}}.
  \bibinfo{pages}{1125--1134}.
\newblock


\bibitem[Khatun et~al\mbox{.}(2021)]%
        {khatun2021systematic}
\bibfield{author}{\bibinfo{person}{Ayesha Khatun},
  \bibinfo{person}{Mohammad~Sajid Shahriar}, \bibinfo{person}{Md~Hasibul
  Hasan}, \bibinfo{person}{Krishna Das}, \bibinfo{person}{Sabbir Ahmed}, {and}
  \bibinfo{person}{Md~Sakibul Islam}.} \bibinfo{year}{2021}\natexlab{}.
\newblock \showarticletitle{A Systematic Review on the Chronological
  Development of Bangla Sign Language Recognition Systems}. In
  \bibinfo{booktitle}{\emph{2021 Joint 10th International Conference on
  Informatics, Electronics \& Vision (ICIEV) and 2021 5th International
  Conference on Imaging, Vision \& Pattern Recognition (icIVPR)}}. IEEE,
  \bibinfo{pages}{1--9}.
\newblock


\bibitem[Kipp et~al\mbox{.}(2011)]%
        {KippAvat}
\bibfield{author}{\bibinfo{person}{Michael Kipp}, \bibinfo{person}{Alexis
  Héloir}, {and} \bibinfo{person}{Quan Nguyen}.}
  \bibinfo{year}{2011}\natexlab{}.
\newblock \showarticletitle{Sign Language Avatars: Animation and
  Comprehensibility}. \bibinfo{pages}{113--126}.
\newblock
\showISBNx{978-3-642-23973-1}
\urldef\tempurl%
\url{https://doi.org/10.1007/978-3-642-23974-8_13}
\showDOI{\tempurl}


\bibitem[Likert(1932)]%
        {likert1932technique}
\bibfield{author}{\bibinfo{person}{Rensis Likert}.}
  \bibinfo{year}{1932}\natexlab{}.
\newblock \showarticletitle{A technique for the measurement of attitudes.}
\newblock \bibinfo{journal}{\emph{Archives of psychology}}
  (\bibinfo{year}{1932}).
\newblock


\bibitem[Miah et~al\mbox{.}(2022)]%
        {miah2022bensignnet}
\bibfield{author}{\bibinfo{person}{Abu Saleh~Musa Miah},
  \bibinfo{person}{Jungpil Shin}, \bibinfo{person}{Md~Al~Mehedi Hasan}, {and}
  \bibinfo{person}{Md~Abdur Rahim}.} \bibinfo{year}{2022}\natexlab{}.
\newblock \showarticletitle{BenSignNet: Bengali Sign Language Alphabet
  Recognition Using Concatenated Segmentation and Convolutional Neural
  Network}.
\newblock \bibinfo{journal}{\emph{Applied Sciences}} \bibinfo{volume}{12},
  \bibinfo{number}{8} (\bibinfo{year}{2022}), \bibinfo{pages}{3933}.
\newblock


\bibitem[Rahim et~al\mbox{.}(2022)]%
        {rahim2022soft}
\bibfield{author}{\bibinfo{person}{Md~Abdur Rahim}, \bibinfo{person}{Jungpil
  Shin}, {and} \bibinfo{person}{Keun~Soo Yun}.}
  \bibinfo{year}{2022}\natexlab{}.
\newblock \showarticletitle{Soft Voting-based Ensemble Model for Bengali Sign
  Gesture Recognition}.
\newblock \bibinfo{journal}{\emph{Annals of Emerging Technologies in Computing
  (AETiC)}} \bibinfo{volume}{6}, \bibinfo{number}{2} (\bibinfo{year}{2022}).
\newblock


\bibitem[Shishir et~al\mbox{.}(2020)]%
        {shishir2020esharagan}
\bibfield{author}{\bibinfo{person}{Fairuz~Shadmani Shishir},
  \bibinfo{person}{Tonmoy Hossain}, {and} \bibinfo{person}{Faisal~Muhammad
  Shah}.} \bibinfo{year}{2020}\natexlab{}.
\newblock \showarticletitle{Esharagan: An approach to generate disentangle
  representation of sign language using infogan}. In
  \bibinfo{booktitle}{\emph{2020 IEEE Region 10 Symposium (TENSYMP)}}. IEEE,
  \bibinfo{pages}{1383--1386}.
\newblock


\bibitem[Syed Tauhid~Ahmed(2016)]%
        {c2}
\bibfield{author}{\bibinfo{person}{M.~A. H.~Akhand Syed Tauhid~Ahmed}.}
  \bibinfo{year}{2016}\natexlab{}.
\newblock \showarticletitle{Bangladeshi Sign Language Recognition using
  Fingertip Position}. In \bibinfo{booktitle}{\emph{International Conference on
  Medical Engineering, Health Informatics and Technology (MediTec)}}.
\newblock


\bibitem[Talukder et~al\mbox{.}(2021)]%
        {talukder2021okkhornama}
\bibfield{author}{\bibinfo{person}{Dipon Talukder}, \bibinfo{person}{Fatima
  Jahara}, \bibinfo{person}{Suvadra Barua}, {and} \bibinfo{person}{Md~Mokammel
  Haque}.} \bibinfo{year}{2021}\natexlab{}.
\newblock \showarticletitle{OkkhorNama: BdSL Image Dataset For Real Time Object
  Detection Algorithms}. In \bibinfo{booktitle}{\emph{2021 IEEE Region 10
  Symposium (TENSYMP)}}. IEEE, \bibinfo{pages}{1--6}.
\newblock


\bibitem[Tang et~al\mbox{.}(2018)]%
        {tang2018gesturegan}
\bibfield{author}{\bibinfo{person}{Hao Tang}, \bibinfo{person}{Wei Wang},
  \bibinfo{person}{Dan Xu}, \bibinfo{person}{Yan Yan}, {and}
  \bibinfo{person}{Nicu Sebe}.} \bibinfo{year}{2018}\natexlab{}.
\newblock \showarticletitle{GestureGAN for Hand Gesture-to-Gesture Translation
  in the Wild}. In \bibinfo{booktitle}{\emph{ACM MM}}.
\newblock


\end{thebibliography}

\appendix

\end{document}